\documentclass[10pt,twocolumn,letterpaper]{article}

\usepackage[pagenumbers]{iccv} 
\usepackage{multirow}
\usepackage{algorithm}
\usepackage{algpseudocode}
\usepackage[table,dvipsnames,xcdraw]{xcolor}
\definecolor{no1}{HTML}{FFB2B2}  
\definecolor{no2}{HTML}{FFD9B2}  
\definecolor{iccvblue}{rgb}{0.21,0.49,0.74}
\usepackage[pagebackref,breaklinks,colorlinks,allcolors=iccvblue]{hyperref}

\def\paperID{11464} 
\def\confName{ICCV}
\def\confYear{2025}

\title{RegGS: Unposed Sparse Views Gaussian Splatting with 3DGS Registration}

\newcommand{\midsize}{\fontsize{11pt}{11pt}\selectfont}
\author{
{\midsize Chong Cheng\textsuperscript{1*}} \and
{\midsize Yu Hu\textsuperscript{1*}} \and
{\midsize Sicheng Yu\textsuperscript{1}} \and
{\midsize Beizhen Zhao\textsuperscript{1}} \and
{\midsize Zijian Wang\textsuperscript{1}} \and
{\midsize Hao Wang\textsuperscript{1\dag}}
\and \midsize \textsuperscript{1}The Hong Kong University of Science and Technology (Guangzhou)
\and {\small \texttt{{ccheng735, yhu847}@connect.hkust-gz.edu.cn}} \quad
{\small \texttt{yusch@mail2.sysu.edu.cn}} \quad
\and {\small \texttt{{bzhao610, zwang886}@connect.hkust-gz.edu.cn}} \quad
{\small \texttt{haowang@hkust-gz.edu.cn}}
}

\begin{document}
\twocolumn[{
\maketitle
\begin{center}
    \vspace{-10pt}
    \includegraphics[width=0.9\textwidth]{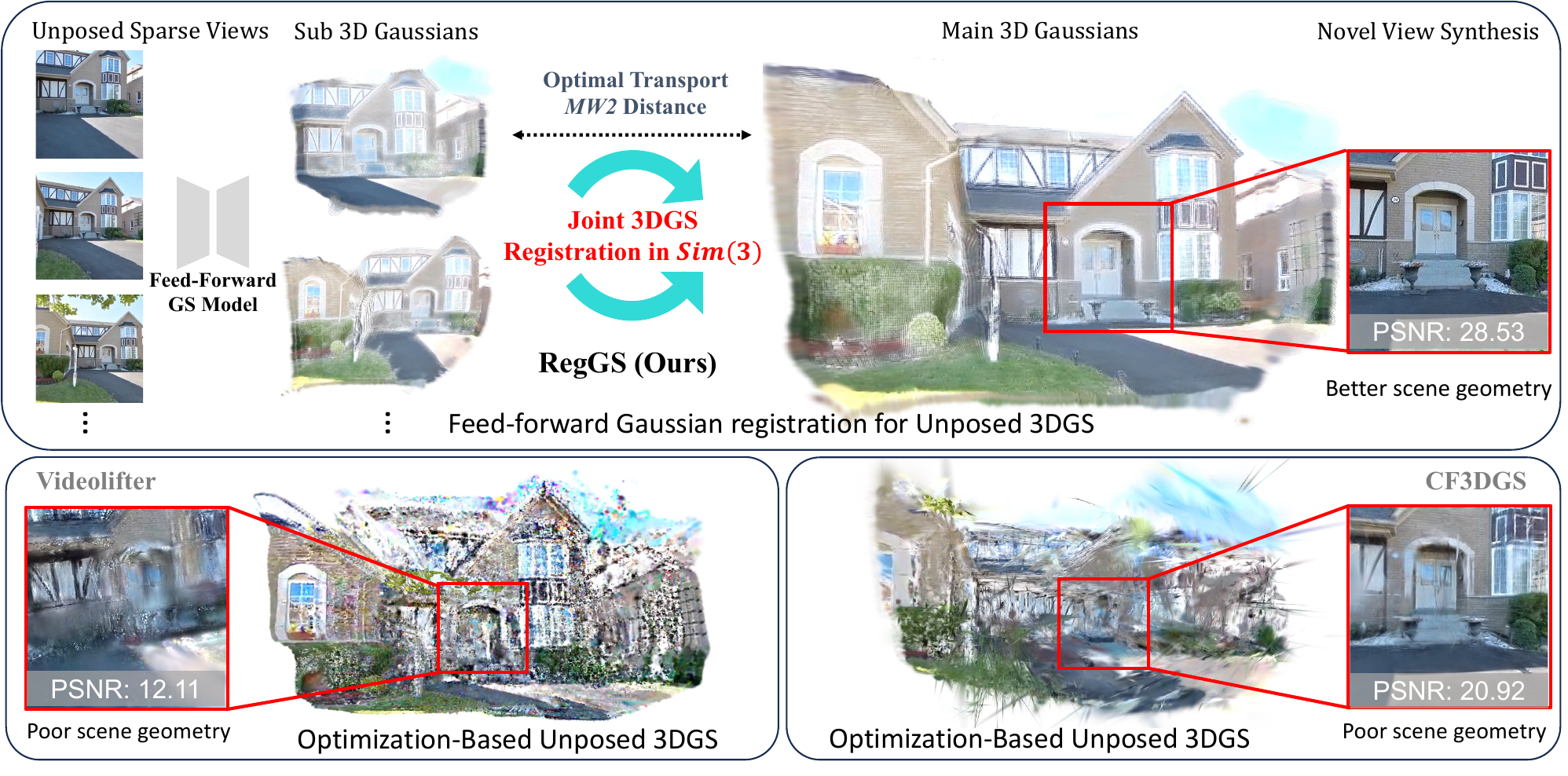}
    \label{fig:teaser}
    \vspace{-10pt}
    \captionof{figure}{
    Overview of our pipeline for 3D Gaussian Splatting from multiple unposed sparse views. A pre-trained feed-forward GS model extracts sub 3D Gaussians from each input, while two initial images yield the main 3D Gaussians. 
    We measure the structural closeness of Gaussian sets using the entropy-regularized $\text{MW}_2$ distance and align them in $\mathrm{Sim}(3)$ space with our joint 3DGS registration module. Our method outperforms others in reconstruction quality and novel view synthesis.
}
\end{center}
}]

\renewcommand{\thefootnote}{}
\footnotetext{* Equal contribution.}
\footnotetext{\dag~Corresponding author.}
\begin{abstract}
3D Gaussian Splatting (3DGS) has demonstrated its potential in reconstructing scenes from unposed images. However, optimization-based 3DGS methods struggle with sparse views due to limited prior knowledge. Meanwhile, feed-forward Gaussian approaches are constrained by input formats, making it challenging to incorporate more input views. To address these challenges, we propose RegGS, a 3D Gaussian registration-based framework for reconstructing unposed sparse views. RegGS aligns local 3D Gaussians generated by a feed-forward network into a globally consistent 3D Gaussian representation. Technically, we implement an entropy-regularized Sinkhorn algorithm to efficiently solve the optimal transport Mixture 2-Wasserstein $(\text{MW}_2)$ distance, which serves as an alignment metric for Gaussian mixture models (GMMs) in $\mathrm{Sim}(3)$ space. Furthermore, we design a joint 3DGS registration module that integrates the $\text{MW}_2$ distance, photometric consistency, and depth geometry. This enables a coarse-to-fine registration process while accurately estimating camera poses and aligning the scene. Experiments on the \textit{RE10K} and \textit{ACID} datasets demonstrate that RegGS effectively registers local Gaussians with high fidelity, achieving precise pose estimation and high-quality novel-view synthesis. Project page: \url{https://3dagentworld.github.io/reggs/}.
\end{abstract}    
\section{Introduction}
\label{sec:introduction}

Recent advances in 3D reconstruction and novel view synthesis—driven by the demand for immersive experiences in VR, AR, and robotics—have yielded impressive results under dense observations \cite{chen2024survey, wu2024recent, fei20243d, dalal2024gaussian, song2024gvkfgaussianvoxelkernel}. Reconstructing 3D scenes from sparse, unposed data remains a formidable challenge, as real-world conditions often provide limited overlap and unreliable camera poses \cite{cheng2025graphguidedscenereconstructionimages}.

Despite the effectiveness of Neural Radiance Fields (NeRF) \cite{mildenhall2021nerf} in novel view synthesis, traditional NeRF methods often require known camera poses \cite{barron2022mip, zhang2020nerf++, liu2020neural, pumarola2021d, tancik2022block}, limiting their broader application. Recent efforts to combine pose estimation with NeRF \cite{bian2023nope, chen2023dbarf, lin2021barf, truong2023sparfneuralradiancefields} face issues of difficult convergence and high computational costs.
Optimization-based 3D Gaussian Splatting (3DGS) \cite{kerbl20233d,ren2024octree, lu2024scaffold, chen2024pgsr,fu2024colmap} methods have shown potential in real-time scene reconstruction but struggle with sparse views due to insufficient geometric priors. These limitations often lead to topological discontinuities and scale ambiguities, significantly reducing their practicality.

In contrast, feedforward-based methods \cite{ye2024no, charatan2024pixelsplat, chen2024mvsplat, hong2024pf3plat, xu2024depthsplat, ziwen2024longlrmlongsequencelargereconstruction} leverage implicit 3D priors learned from large-scale training data, enabling direct prediction of coherent 3D Gaussians from images without iterative optimization. This learned prior not only enhances cross-dataset generalization but also regularizes the reconstruction in scenarios with under-constrained geometric information \cite{yu2025rgbonlygaussiansplattingslam,cheng2025outdoormonocularslamglobal}. Recent approaches \cite{ye2024no,smart2024splatt3rzeroshotgaussiansplatting} achieve direct inference of 3D Gaussian representations from unposed images, eliminating the need for iterative optimization.

However, feed-forward methods can only handle a limited number of input images, restricting their applicability to broader scenarios. This raises an intriguing question: \emph{Can we register locally generated Gaussian models from a feed-forward network into a globally consistent 3D Gaussian representation?}

To address this issue, we propose a novel 3D Gaussian reconstruction framework: \textbf{RegGS}, which performs unposed sparse view reconstruction by registering feed-forward Gaussian incrementally. Specifically, we introduce the optimal transport-based Mixture 2-Wasserstein $(MW2)$ distance between Gaussian mixture models (GMM) to align generalized Gaussian manifolds. Through a differentiable multi-modal joint registration pipeline, we solve for scene alignment in the $\mathrm{Sim(3)}$ space.

Technically, we utilize the entropy-regularized Sinkhorn algorithm to compute the differentiable upper bound $MW2$ for the $W2$ distance between GMMs, thereby circumventing the infinite-dimensional $W2$ optimization problem.
By integrating engineering techniques such as log-Sinkhorn and Cholesky decomposition, we efficiently compute the $MW2$ distance between thousands of 3D Gaussians on GPU, thereby accurately measuring their alignment in the $\mathrm{Sim(3)}$ space.

Furthermore, we incorporate the global distribution of the MW2 distance, photometric consistency, and depth geometry into a joint 3D Gaussian registration module, enabling elastic scale alignment and topology adaptation within $\mathrm{Sim(3)}$. By performing a coarse-to-fine incremental 3DGS registration followed by global optimization, we achieve high-precision camera pose estimation and high-quality scene reconstruction. Our contribution can be summarized as:
\begin{itemize}
\item We construct an optimal transport framework for Gaussian Mixture Models in the $\mathrm{Sim(3)}$ space and efficiently compute the $MW_2$ distance using the entropy-regularized Sinkhorn algorithm, thereby providing a differentiable alignment metric for 3D Gaussian distributions.

\item We propose a 3DGS joint registration module that achieves precise camera pose estimation and scene registration by jointly utilizing MW2 distance, photometric consistency, and depth geometry.

\item Experiments on the RE10K and ACID datasets demonstrate that
RegGS significantly improves pose estimation accuracy and the quality of novel view synthesis, offering broad possibilities for practical applications.
\end{itemize}
\section{Related Work}
\label{sec:related}

\subsection{NeRF-based Pose-Free Reconstruction}

Novel view synthesis, particularly in the absence of accurate camera poses, has garnered significant attention in recent years.
Traditional Neural Radiance Fields (NeRF) methods~\cite{mildenhall2021nerf, barron2022mip, zhang2020nerf++, liu2020neural} have achieved remarkable results. 
However, these methods usually rely on known camera poses for training, limiting their applicability in scenarios where pose information is unavailable or unreliable which is very common in real-world scenarios. 

Several approaches have been proposed to extend NeRF to handle unposed input images.
Among them, \cite{chen2023dbarf, cheng2023lu} integrate camera pose estimation with NeRF rendering, leveraging a recurrent GRU module for pose and depth estimation.
Similarly, \cite{smith2023flowcam} employs a weighted Procrustes analysis and an optical flow network to establish correspondences for pose estimation. 
More recently, CoPoNeRF~\cite{hong2024unifying} introduced a unified framework that integrates correspondence matching, pose estimation, and NeRF rendering, allowing for end-to-end training and improved performance in challenging scenarios with extreme viewpoint changes.
Additionally, methods like Nope-NeRF~\cite{bian2023nope} leverage depth information to constrain the optimization process. 

While NeRF-based methods show promise, their reliance on dense ray sampling leads to slow training and inference, struggles with extreme viewpoint changes and minimal overlap, and high computational costs.

\subsection{Optimization-based Pose-Free 3DGS Reconstruction}

3D Gaussian Splatting (3DGS)~\cite{kerbl20233d} offers an alternative by representing the scene with a set of 3D Gaussians, which can be rendered efficiently.
However, traditional 3DGS also relies on accurate camera poses and sparse point clouds from Structure-from-Motion (SfM) pipelines like Colmap.

To address this, Colmap-Free 3DGS~\cite{fu2024colmap} proposes a method to optimize the 3D Gaussian representation directly from unposed images. 
By incorporating pose estimation into the optimization loop, this approach eliminates the need for precomputed poses, making it more flexible and applicable to a wider range of scenarios. 

Similarly, videoLifter uses pre-trained models~\cite{leroy2024groundingimagematching3d, Wang_2024_CVPR} to reconstruct globally consistent 3D models from uncalibrated monocular videos, reducing error accumulation and computational costs. Yet, it struggles with sparse view reconstruction challenges. While optimization-based methods can achieve high-quality reconstructions, they struggle to efficiently handle sparse viewpoint scenes and face challenges in learning complex 3D spatial relationships.

\subsection{Feedforward-based Pose-Free 3DGS Reconstruction}
Feed-forward approaches aim to alleviate this by predicting the 3D representation directly from the input images in a single pass. 
NoPoSplat~\cite{ye2024no} exemplifies this by using a neural network to map unposed images to a 3D Gaussian representation in a canonical space, enabling fast and efficient reconstruction without iterative optimization. 
Other feed-forward methods, such as pixelSplat~\cite{charatan2024pixelsplat} and MVSplat~\cite{chen2024mvsplat}, predict Gaussian primitives from posed images, leveraging geometric priors like epipolar geometry or cost volumes. 

In contrast, NoPoSplat operates without poses by directly predicting Gaussians in a canonical space, demonstrating improved performance, especially in scenarios with limited overlap between input views.
However, feed-forward Gaussian models typically handle only a limited number of input images, limiting their application in scenarios with large coverage and sparse viewpoints.

Consequently, we explored a method based on 3D Gaussian registration to achieve incremental unposed sparse view reconstruction. This approach not only leverages the excellent scene priors of feed-forward models but also enables high-quality reconstruction in broader sparse view scenarios, which is of practical importance.
\section{Method}
\begin{figure*}
  \centering
  \includegraphics[width=\textwidth]{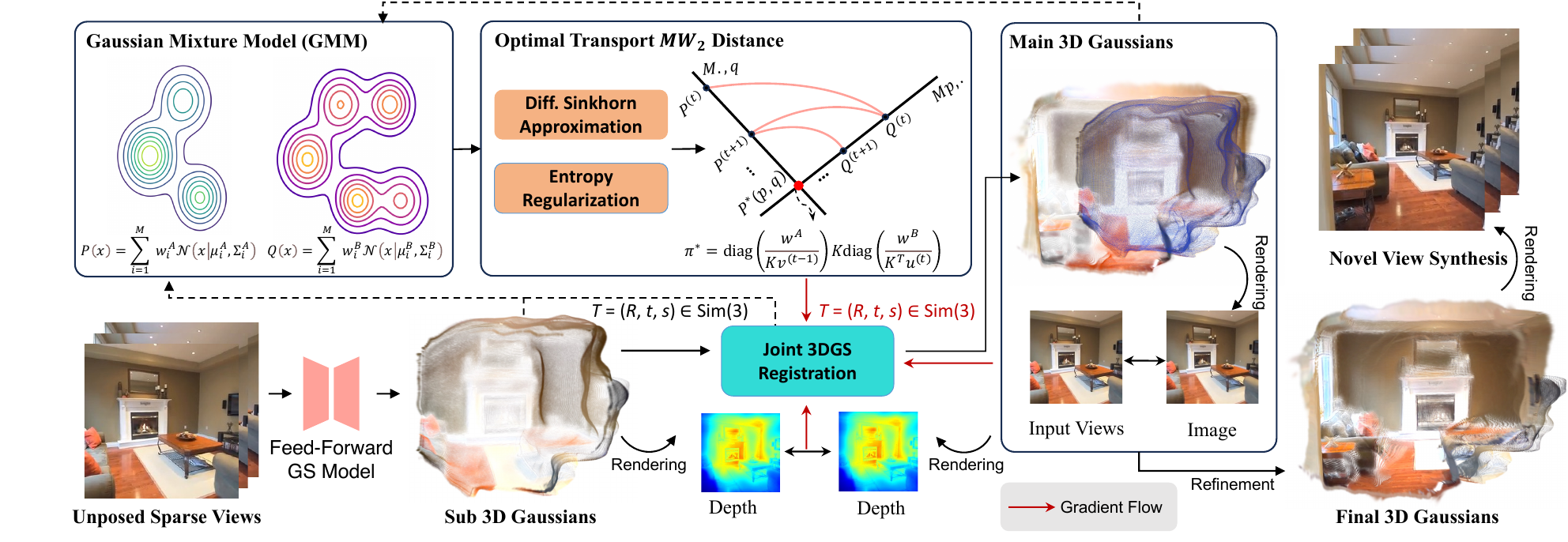}
  \vspace{-15pt}
  \caption{\textbf{Pipeline of Unposed Sparse Views Gaussian Splatting with 3DGS Registration (RegGS).} First, we use a pre-trained feed-forward Gaussian model to construct a main Gaussians from two initial images. Then, for each new input, a sub Gaussians is generated and aligned with the main Gaussians. Specifically, by solving the optimal transport $MW2$ distance with an entropy-regularized Sinkhorn approximation, our differentiable 3DGS joint registration module estimates the $\mathrm{Sim}(3)$ transformation and merges the sub Gaussians into the main Gaussians. Finally, we perform refinement of the global Gaussians, yielding a high-fidelity 3D reconstruction.
  }
  \label{fig:framework}
\end{figure*}

As shown in \cref{fig:framework}, our method initializes a main map from two images using a pretrained feed-forward Gaussian model, and generates sub Gaussians for each subsequent image. By measuring similarity between the GMMs through an optimal transport $MW2$ distance by an entropy-regularized Sinkhorn approach, our differentiable joint 3DGS registration module estimates the Sim(3) transformation before merging local Gaussians into the main map. Finally, we perform a global refinement of the 3D Gaussians with adaptive pruning, yielding high-fidelity reconstructions even from unposed sparse views.

\subsection{Registration Problem Modeling}
The core of our work is 3DGS registration. An intuitive approach is to use 3DGS center points as registration references. However, these center points cannot accurately reflect the geometric structure of the scene. 
Here, we introduce a statistical model, Gaussian Mixture Model (GMM) \cite{doi:10.1137/19M1301047}, which can describe the structural distribution of 3D Gaussians based on their attributes. Specifically, we first define the main 3D Gaussians between two frames, with the main Gaussians $\mathcal{G}^A$ and sub Gaussians $\mathcal{G}^B$ expressed as GMMs:
\begin{equation}
\small
G^A = \sum_{i=1}^{M} w_i^A \mathcal{N}(\mu_i^A, \Sigma_i^A),
\end{equation}
\begin{equation}
\small
G^B = \sum_{k=1}^{N} w_k^B \mathcal{N}(\mu_k^B, \Sigma_k^B),
\end{equation}
where $\mu$ represents the mean of the Gaussian distribution, $\Sigma$ represents the covariance matrix, and weights satisfy $\sum_i w_i^A = 1$, $\sum_k w_k^B = 1$, obtained through opacity normalization. 

It is notable that we do not consider color information (spherical harmonic coefficients), as color information is unstable due to lighting angle variations. Our goal is to find the optimal \textbf{affine transformation} $T \in \mathrm{Sim}(3)$ parameters, including rotation $R \in SO(3)$, translation $t \in \mathbb{R}^3$, and scaling factor $s \in \mathbb{R}^+$, such that the structural difference between the transformed sub Gaussians $T(\mathcal{G}^B)$ and the main Gaussians $\mathcal{G}^A$ is minimized. The objective function is:
\begin{equation}
\small
T^* = \arg\min_{T \in \text{Sim}(3)} \mathcal{D}\left( \mathcal{G}^A, T(\mathcal{G}^B) \right),
\end{equation}
where $\mathcal{D}$ is a distance metric function used to measure the difference between two sets of 3D Gaussian distributions. After the Sim(3) transformation of the sub-map, the parameters of each Gaussian component change according to the following relationships:
\begin{equation}
\small
\mu_k^{B'} = s\,R\,\mu_k^B + t,\qquad
\Sigma_k^{B'} = s^2\,R\,\Sigma_k^B\,R^\top.
\end{equation}
Under the above transformation, we compute the matching relationship between Gaussian components in the main Gaussians $\mathcal{G}^A$ and the transformed sub Gaussians $T(\mathcal{G}^B)$ by minimizing the $\mathcal{D}$ distance.

\subsection{Optimal Transport $MW_2$ Distance}
\label{sec:sinkhorn}
Inspired by previous research \cite{JMLR:v22:20-588}, we adopt the 2-Wasserstein $(W_2)$ distance as the fundamental metric to measure geometric differences between two sets of 3D Gaussian distributions. For two Gaussian components $\mathcal{N}(\mu_i^A, \Sigma_i^A)$ and $\mathcal{N}(\mu_k^{B'}, \Sigma_k^{B'})$, the square of their $W2$ distance is defined as:
\begin{equation}
\small
W_2^2 = |\mu_i^A - \mu_k^{B'}|^2 + \text{Tr}\left( \Sigma_i^A + \Sigma_k^{B'} - 2\left( \Sigma_i^A \Sigma_k^{B'} \right)^{1/2} \right),
\end{equation}
where the position term $\|\mu_i^A - \mu_k^{B'}\|^2$ reflects the Euclidean offset between distribution centers, and the covariance term eliminates rotation effects through matrix square roots, becoming zero when $\Sigma_i^A = \Sigma_k^{B'}$. 

However, directly computing the $W_2$ distance between GMMs requires solving an infinite-dimensional optimization problem, which is computationally infeasible \cite{Altschuler_2022}. To address this, we introduce the ``GMM transport" method, which constrains the optimal transport plan to the Gaussian mixture subspace, transforming the continuous problem into a discrete linear assignment problem \cite{doi:10.1137/19M1301047}. Its mathematical form is:
\begin{equation}
\small
\mathrm{MW}_2^2(P,Q) = \inf_{\pi \in \Pi(w^A,w^B)} \sum_{i=1}^M \sum_{k=1}^N \pi_{ik} C_{ik},
\end{equation}
where $C_{ik}$ is the transport cost for the Gaussian pair $(i,k)$, and $\Pi(w^A,w^B)$ is the set of transport plans satisfying $\sum_i \pi_{ik} = w_k^B$ and $\sum_k \pi_{ik} = w_i^A$. In this case, $\mathrm{MW}_2$ forms an upper bound of $W_2$, satisfying $\mathrm{MW}_2(\mu_0,\mu_1) \geq W_2(\mu_0,\mu_1)$ \cite{doi:10.1137/19M1301047}.

We employ the optimal transport Sinkhorn algorithm \cite{Sinkhorn} to compute the $MW_2$ distance. Since the two sets of Gaussian spheres are not in one-to-one correspondence and are numerous, to avoid local minima, accelerate convergence, and enable fuzzy matching, we employ an entropy regularization strategy to construct a differentiable Sinkhorn approximation. The optimization objective is:

\begin{equation}
\small
W_{2,\epsilon}^2 = \min_{\pi \in \Pi(w^A,w^B)} \left[ \sum_{i,k} \pi_{ik} C_{ik} + \epsilon \sum_{i,k} \pi_{ik} \log \pi_{ik} \right],
\end{equation}

where $\epsilon$ controls the regularization strength. We solve this problem through Sinkhorn iterations: initially, we initialize the kernel matrix $K_{ik} = \exp(-C_{ik}/\epsilon)$; subsequently, we alternately perform scaling updates:
\begin{equation}
\small
u^{(t)} = \frac{w^A}{K v^{(t-1)}}, \quad v^{(t)} = \frac{w^B}{K^\top u^{(t)}}.
\end{equation}
After $T$ iterations, we obtain the transport plan 
\begin{equation}
\small
\pi^* = \text{diag}(u^{(T)}) K \text{diag}(v^{(T)}),
\end{equation}
and finally calculate the entropy-regularized Wasserstein distance 
\begin{equation}
\small
W_{2,\epsilon}^2 = \sum_{i,k} \pi_{ik}^* C_{ik}.
\end{equation}
This method reduces the computational complexity to $O(MN)$ while ensuring gradient differentiability. The proof of gradient consistency for the entropy-regularized Sinkhorn $W_2$ distance, along with the complexity calculations, can be found in the appendix.

\subsection{Differentiable Joint 3DGS Registration}
\label{sec:joint_registration}
To establish an efficient and stable 3D Gaussian registration model, we propose a differentiable framework based on quaternion parameterization and multi-objective joint optimization. In traditional methods, pose parameterization often faces redundancy or singularity issues, and our proposed Sinkhorn approximation of $MW_2$ distance is not an exact solution, making single-objective optimization prone to local optima. Therefore, we design a strategy that integrates quaternion pose representation, multi-loss joint optimization, and adaptive weight allocation, with mathematical formulation and implementation details as follows.

\noindent\textbf{Pose Parameterization Design}: We represent a $\mathrm{Sim}(3)$ transformation by decomposing it into a quaternion rotation $\mathbf{q}\in S^3$, a translation $\mathbf{t}\in\mathbb{R}^3$, and a logarithmic scale $\log s\in\mathbb{R}$, forming the parameter vector $\boldsymbol{\theta} = [\mathbf{q}; \mathbf{t}; \log s] \in \mathbb{R}^8$. This formulation guarantees positive scaling via $s = \exp(\log s)$ and enforces $\|\mathbf{q}\| = 1$ using projected gradient updates. When applied to Gaussian components, the update formulas for mean and covariance are:
\begin{equation}
\begin{aligned}
\small
\mu_k^{B'} &= s \cdot R(\mathbf{q}) \mu_k^B + \mathbf{t}, \\
\Sigma_k^{B'} &= s^2 \cdot R(\mathbf{q}) \Sigma_k^B R(\mathbf{q})^\top,
\end{aligned}
\end{equation}
where the rotation matrix $R(\mathbf{q})$ is analytically generated from the quaternion $\mathbf{q} = [w, x, y, z]^\top$:
\begin{equation}
\setlength{\arraycolsep}{1pt}  
\small
R(\mathbf{q}) = \begin{bmatrix}
1 - 2y^2 - 2z^2 & 2xy - 2wz & 2xz + 2wy \\
2xy + 2wz & 1 - 2x^2 - 2z^2 & 2yz - 2wx \\
2xz - 2wy & 2yz + 2wx & 1 - 2x^2 - 2y^2
\end{bmatrix}.
\end{equation}

Our experiments show that quaternion rotation converges significantly faster than Lie algebra rotation while achieving equivalent accuracy.

\noindent\textbf{Multi-Loss Joint Optimization}: To balance global distribution alignment and precise geometric consistency, we construct a joint loss function:
\begin{equation}
\mathcal{L}_{\text{total}} = \lambda_1 \mathcal{L}_{\text{MW}_2} + \lambda_2 \mathcal{L}_{\text{Photo}} + \lambda_3 \mathcal{L}_{\text{Depth}},
\end{equation}
where the global alignment term $\mathcal{L}_{\text{MW}_2} = W_{2,\epsilon}^2(G^A, T(G^B))$ is calculated using the differentiable Sinkhorn algorithm from \cref{sec:sinkhorn}, driving the overall matching of Gaussian distribution centers and covariances; the local photometric term uses the 3DGS differentiable rendering pipeline \cite{kerbl20233d} to generate RGB images from aligned viewpoints, enhancing precise map alignment through pixel-level L1 loss. The local photometric loss is described as:
\begin{equation}
\small
\mathcal{L}_{\text{Photo}} = \frac{1}{|P|} \sum_{p \in P} \left| I^A(p) - I^{T(B)}(p) \right|_1,
\end{equation}
where $T(G^B)$ represents applying the current Sim(3) transformation with parameters $\boldsymbol{\theta}$ to the source distribution $G^B$; depth is similarly rendered using the 3DGS differentiable rendering pipeline \cite{kerbl20233d}, with invalid regions excluded through an effective depth mask $M_v$, suppressing scale drift and topological distortion. The depth geometric constraint term is described as:
\begin{equation}
\small
\mathcal{L}_{\text{Depth}} = \frac{1}{|M_v|} \sum_{p \in M_v} \left| D_A^v(p) - D_{T(B)}^v(p) \right|,
\end{equation}
where $D_A^v(p) \in \mathbb{R}^+$ and $D_{T(B)}^v(p) \in \mathbb{R}^+$ are depth maps under viewpoint $v$, and $M_v$ is the valid depth mask.

\noindent\textbf{Differentiable Gradient Path}: To achieve end-to-end optimization, we calculate the gradient of the loss with respect to parameters $\boldsymbol{\theta}$. For the $MW_2$ term, its gradient propagates through the transport plan $\pi_{ik}^*$ and the chain rule:
\begin{equation}
\small
\frac{\partial \mathcal{L}_{\text{MW}_2}}{\partial \boldsymbol{\theta}} = \sum_{i,k} \pi_{ik}^* \left( \frac{\partial C_{ik}}{\partial \mu_k^{B'}} \frac{\partial \mu_k^{B'}}{\partial \boldsymbol{\theta}} + \frac{\partial C_{ik}}{\partial \Sigma_k^{B'}} \frac{\partial \Sigma_k^{B'}}{\partial \boldsymbol{\theta}} \right),
\end{equation}
where the Jacobian matrix of quaternion rotation $\partial R(\mathbf{q})/\partial \mathbf{q}$ is implicitly solved by automatic differentiation. The gradients of photometric and depth terms are back-propagated through the rendering pipeline:
\begin{equation}
\small
\frac{\partial \mathcal{L}_{\text{Photo}}}{\partial \boldsymbol{\theta}} = \frac{1}{|P|} \sum_p \text{sign}(I^A - I^{T(B)}) \cdot \frac{\partial I^{T(B)}}{\partial \mu_k^{B'}} \frac{\partial \mu_k^{B'}}{\partial \boldsymbol{\theta}},
\end{equation}
\begin{equation}
\small
\frac{\partial \mathcal{L}_{\text{Depth}}}{\partial \boldsymbol{\theta}} = \frac{1}{|M_v|} \sum_{p \in M_v} \text{sign}(D_A^v - D_{T(B)}^v) \cdot \frac{\partial D_{T(B)}^v}{\partial \mu_k^{B'}} \frac{\partial \mu_k^{B'}}{\partial \boldsymbol{\theta}},
\end{equation}
where the rendering gradients $\partial I/\partial \mu_k^{B'}$ and $\partial D/\partial \mu_k^{B'}$ are analytically derived from the 3DGS volume rendering formula \cite{kerbl20233d}.

The joint optimization of these three components allows for fast and robust registration of 3DGS sub-maps. Subsequently, the next frame is inferred as a sub-map by the pre-trained model, continuously updating the main map to complete the reconstruction.

\begin{figure*}
  \centering
  \includegraphics[width=0.92\textwidth]{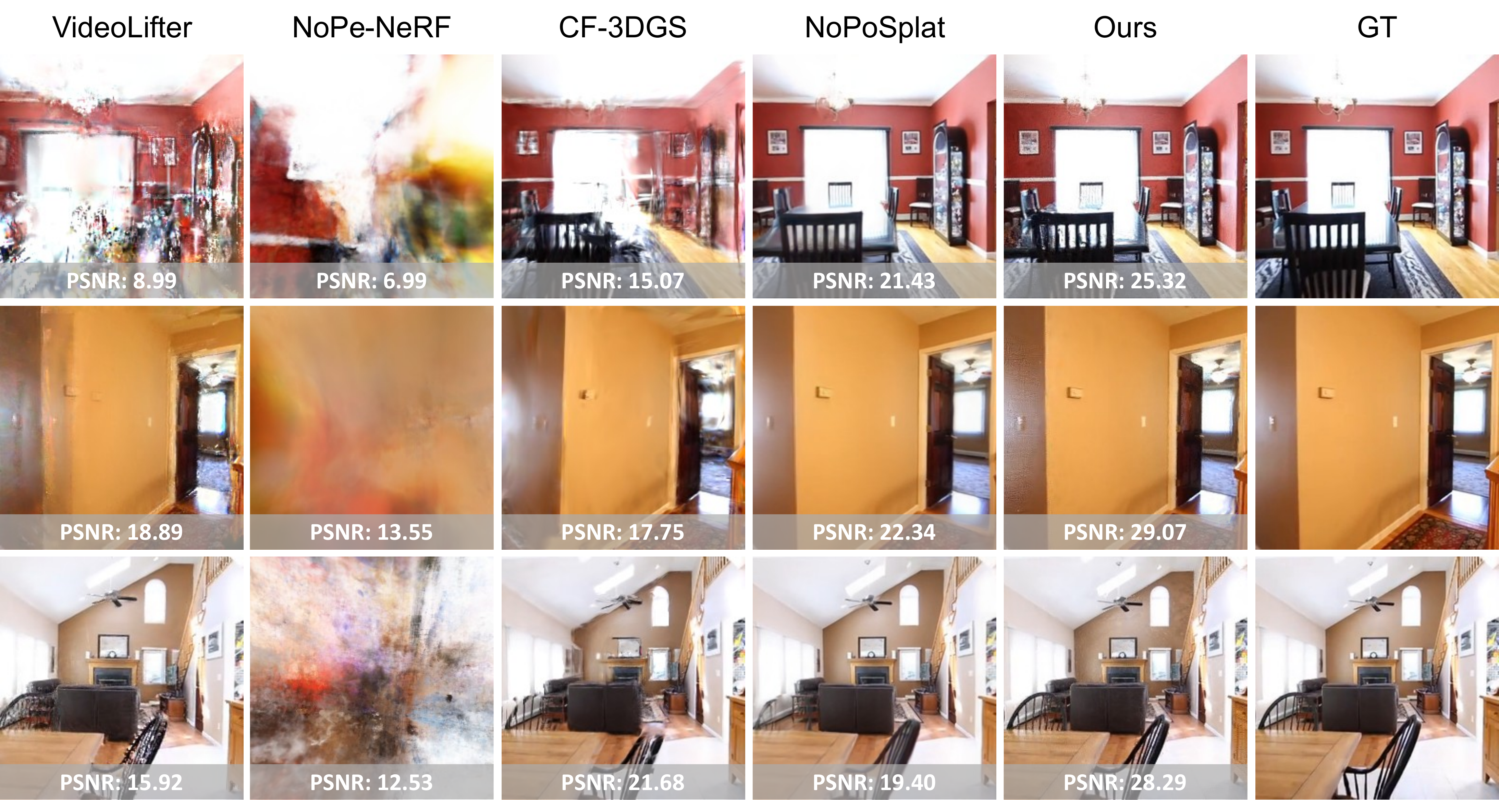}
   \vspace{-3pt}
  \caption{\textbf{Qualitative Comparison on the RE10K \cite{zhou2018stereo}.} NoPoSplat: 2$\times$ views; others: 16$\times$ views. Our method not only registers the 3D Gaussians but also enhances novel view synthesis through global refinement.}
  \label{fig:re10k_visual_diff}
  \vspace{-3pt}
\end{figure*}

\begin{table*}
\small
\centering
\setlength{\tabcolsep}{3pt}
\begin{tabular}{@{}l|ccc|ccc|ccc|ccc@{}}
\toprule
                          \multirow{2}{*}{\textbf{Method}} & \multicolumn{3}{c}{2×}   & \multicolumn{3}{c}{8×}   & \multicolumn{3}{c}{16×}  & \multicolumn{3}{c}{32×} \\ \cmidrule(lr){2-4} \cmidrule(lr){5-7} \cmidrule(lr){8-10} \cmidrule(lr){11-13}  
& PSNR↑  & SSIM↑  & LPIPS↓ & PSNR↑  & SSIM↑  & LPIPS↓ & PSNR↑  & SSIM↑  & LPIPS↓ & PSNR↑  & SSIM↑  & LPIPS↓ \\ \hline
COLMAP* \cite{schoenberger2016sfm} & 9.687 & 0.266 & 0.533   & 7.171 & 0.135 & 0.676   & 18.904 & 0.614 & 0.294   & 22.911 & 0.725 & 0.219 \\
Splatt3R \cite{smart2024splatt3rzeroshotgaussiansplatting} & 13.951 & 0.442 & 0.443   & - & - & - & - & - & - & - & - & - \\
NoPoSplat \cite{ye2024no} &\cellcolor{no2}23.247 & \cellcolor{no2}0.832 & \cellcolor{no1}0.111   & - & - & - & - & - & - & - & - & - \\
CF-3DGS \cite{fu2024colmap} & 19.326 & 0.638 & 0.277   & 20.329 & 0.672 & 0.235   & 23.034 & 0.792 & 0.188   & 25.596 & 0.865 & \cellcolor{no2} 0.133 \\
NoPeNerf \cite{bian2023nope} & 10.225 & 0.351 & 0.781   & 10.974 & 0.343 & 0.767   & 10.465 & 0.321 & 0.763   & 10.021 & 0.284 & 0.742 \\
VideoLifter \cite{cong2025videolifter} & 14.526 & 0.448 & 0.346   & 16.651 & 0.564 & 0.273   & 14.765 & 0.452 & 0.382   & 15.268 & 0.483 & 0.344 \\
MASt3R* \cite{leroy2024groundingimagematching3d} & 16.036 & 0.580 & 0.361   & \cellcolor{no2}24.249 & \cellcolor{no2}0.824 & \cellcolor{no2}0.189   & \cellcolor{no2}27.024 & \cellcolor{no2}0.869 & \cellcolor{no2}0.149   & \cellcolor{no2}28.309 & \cellcolor{no2}0.891 & \cellcolor{no1}0.094 \\
\hline
\textbf{RegGS (Ours)} & \cellcolor{no1}\textbf{24.272} & \cellcolor{no1}\textbf{0.853} & \cellcolor{no2}\textbf{0.174}   & \cellcolor{no1}\textbf{26.691} & \cellcolor{no1}\textbf{0.877} & \cellcolor{no1}\textbf{0.185}   & \cellcolor{no1}\textbf{28.663} & \cellcolor{no1}\textbf{0.913} & \cellcolor{no1}\textbf{0.147}   & \cellcolor{no1}\textbf{28.332} & \cellcolor{no1}\textbf{0.912} & 0.151 \\
\bottomrule
\end{tabular}
\vspace{-3pt}
\caption{\textbf{Novel View Synthesis Results on the RE10K \cite{zhou2018stereo}.} The terms ``2x'', ``8x'', ``16x'', and ``32x'' represent the number of views in the input images. An asterisk (*) indicates reconstruction with 3DGS. A dash (-) indicates that the input is not supported by the method. Our method outperforms other unposed methods in reconstruction quality with sparse views, and the gap widens as the number of views decreases.}
\label{tab:re10k_comparison}
\vspace{-3pt}
\end{table*}

\subsection{Joint Training}
\noindent\textbf{Joint 3DGS Registration.} Feed-forward Gaussian models often produce targets with vastly different scales. To avoid falling into local optima, we perform scale normalization before optimization. We begin by calculating the average value of depth rendered from sub Gaussian map, which is generated by the feed-forward Gaussian model, denoted as \( D_{\text{sub}} \), and scale it to a common scale. Moreover, in joint optimization, to enhance the efficiency of iterative optimization, initialization is also necessary. We compare the depth values of the main Gaussians function \( D_{\text{main}} \) with those of the sub Gaussians function \( D_{\text{sub}} \) to determine the initial relative scale \( \text{s}_{\text{init}} \). 

\noindent\textbf{Computational Efficiency.} To achieve efficient computation of large-scale Gaussian $\text{MW}_2$ distances, we map Sinkhorn iteration operations, including matrix scaling, covariance matrix Cholesky decomposition, and Wasserstein distance calculation to GPU through tensorized operations, achieving efficient computation between Gaussian pairs through batch parallel processing. To address the risk of exponential term overflow in entropy regularization, we design a logarithmic space accumulation strategy that maintains numerical stability when computing $\text{MW}_2$, while uniformly regularizing covariance matrices as $\Sigma \leftarrow \Sigma + 10^{-6}I$ to ensure positive definiteness.

\begin{figure*}
  \centering
  \includegraphics[width=0.92\textwidth]{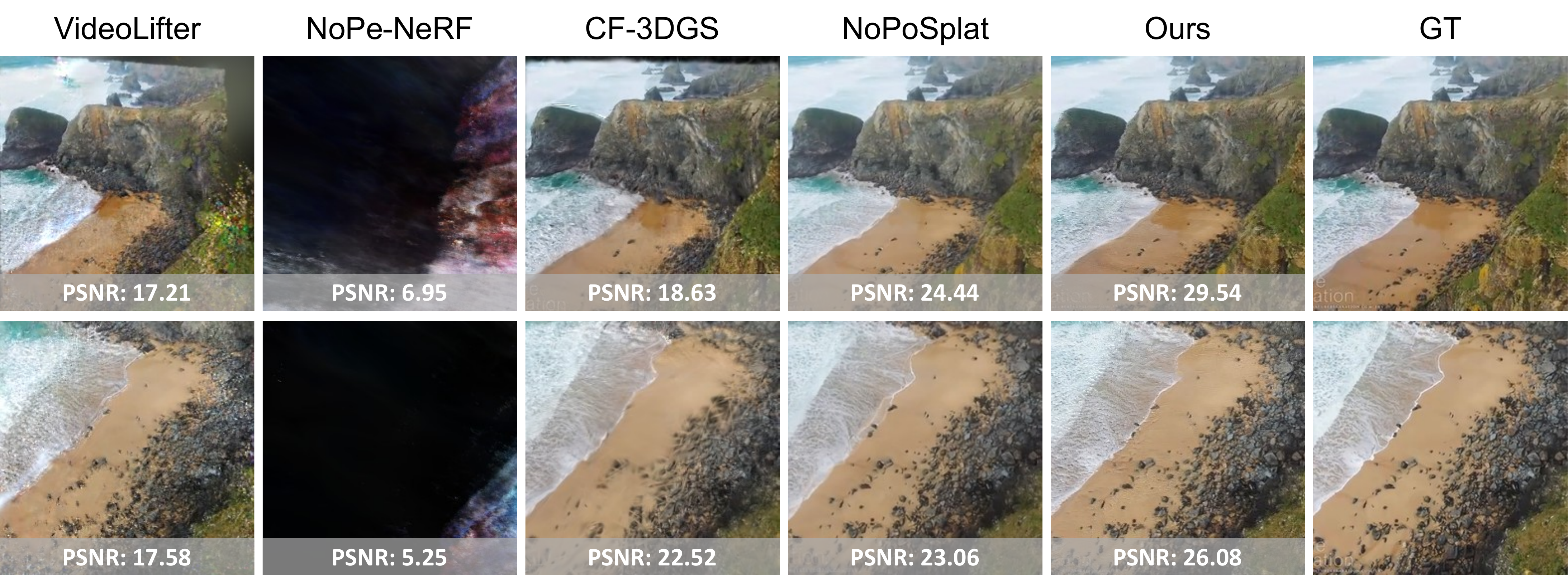}
  \vspace{-3pt}
  \caption{\textbf{Qualitative comparison on the ACID \cite{infinite_nature_2020}.}  NoPoSplat: 2x views; others: 16x views. Our method is applicable to both indoor scenes and drone-captured videos, demonstrating superior novel view synthesis performance.}
  \label{fig:acid_visual_diff}
  \vspace{-3pt}
\end{figure*}

\begin{table*}[]
\small
\centering
\setlength{\tabcolsep}{3pt}

\begin{tabular}{@{}l|ccc|ccc|ccc|ccc@{}}
\toprule
                          \multirow{2}{*}{\textbf{Method}} & \multicolumn{3}{c}{2×} & \multicolumn{3}{c}{8×} & \multicolumn{3}{c}{16×} & \multicolumn{3}{c}{32×} \\ \cmidrule(lr){2-4} \cmidrule(lr){5-7} \cmidrule(lr){8-10} \cmidrule(lr){11-13} 
& PSNR↑  & SSIM↑  & LPIPS↓ & PSNR↑  & SSIM↑  & LPIPS↓ & PSNR↑  & SSIM↑  & LPIPS↓ & PSNR↑  & SSIM↑  & LPIPS↓ \\ \hline
COLMAP* \cite{schoenberger2016sfm} & 8.340 & 0.141 & 0.643   & 14.162 & 0.207 & 0.554   & 7.904 & 0.049 & 0.719   & 7.300 & 0.058 & 0.716 \\
Splatt3R \cite{smart2024splatt3rzeroshotgaussiansplatting} & 10.468 & 0.215 & 0.591   & - & - & - & - & - & - & - & - & - \\
NoPoSplat \cite{ye2024no} & \cellcolor{no2}23.589 & \cellcolor{no2}0.663 & \cellcolor{no1}0.202   & - & - & - & - & - & - & - & - & - \\
CF-3DGS \cite{fu2024colmap} & 21.654 & 0.604 & 0.301   & 22.212 & \cellcolor{no2}0.629 & \cellcolor{no2}0.289   & 23.458 & 0.651 & 0.266   & 23.419 & 0.650 & 0.263 \\
NoPeNerf \cite{bian2023nope} & 13.231 & 0.269 & 0.748   & 14.611 & 0.273 & 0.732   & 6.837 & 0.117 & 0.788   & 11.961 & 0.222 & 0.756 \\
VideoLifter \cite{cong2025videolifter} & 17.921 & 0.327 & 0.405   & 18.830 & 0.332 & 0.394   & 18.264 & 0.289 & 0.412   & 19.503 & 0.393 & 0.335 \\
MASt3R* \cite{leroy2024groundingimagematching3d} & 18.390 & 0.312 & 0.447   & \cellcolor{no2}22.231 & 0.525 & 0.318   & \cellcolor{no2}24.537 & \cellcolor{no2}0.673 & \cellcolor{no2}0.240   & \cellcolor{no2}25.216 & \cellcolor{no2}0.702 & \cellcolor{no1}0.155 \\
\hline
\textbf{RegGS (Ours)} & \cellcolor{no1}\textbf{24.291} & \cellcolor{no1}\textbf{0.703} & \cellcolor{no2}\textbf{0.237}   & \cellcolor{no1}\textbf{25.764} & \cellcolor{no1}\textbf{0.753} & \cellcolor{no1}\textbf{0.252}   & \cellcolor{no1}\textbf{27.745} & \cellcolor{no1}\textbf{0.834} & \cellcolor{no1}\textbf{0.201}   & \cellcolor{no1}\textbf{26.772} & \cellcolor{no1}\textbf{0.774} & \cellcolor{no2}\textbf{0.243} \\
\bottomrule
\end{tabular}
\vspace{-3pt}
\caption{\textbf{Novel View Synthesis Results on the ACID \cite{infinite_nature_2020}.} The terms ``2x'', ``8x'', ``16x'', and ``32x'' represent the number of views in the input images. An asterisk (*) indicates reconstruction with 3DGS. A dash (-) indicates that the input is not supported by the method.  The data shows that our method also outperforms other unposed reconstruction methods in drone-captured scenes. As the scene becomes sparser, the gap between our method and the others increases.}
\label{tab:acid_comparison}
\vspace{-3pt}
\end{table*}

\begin{table}[]
\scriptsize
\centering
\setlength{\tabcolsep}{2pt} 
\begin{tabular}{l|ccc|ccc}
\toprule
      \multirow{2}{*}{\textbf{Method}} &        & RE10K   &         &        & ACID    &         \\
 & 8x ATE↓ & 16x ATE↓ & 32x ATE↓ & 8x ATE↓ & 16x ATE↓ & 32x ATE↓ \\ \hline
VideoLifter                & 0.335 & 0.291             &\cellcolor{no2}0.232 & \cellcolor{no2}0.272 & 0.206             & \cellcolor{no2}0.145 \\
NoPeNerf                   & 0.844             & 0.902             & 0.597             & 0.684             & 0.413             & 0.455             \\
CF3DGS                     & \cellcolor{no2}0.237                & \cellcolor{no2}0.254 & 0.286             & 0.278                 & \cellcolor{no2}0.195 & 0.239             \\ \hline
Ours                       & \cellcolor{no1}\textbf{0.023}    & \cellcolor{no1}\textbf{0.041}    & \cellcolor{no1}\textbf{0.078}    & \cellcolor{no1}\textbf{0.020}    & \cellcolor{no1}\textbf{0.038}    & \cellcolor{no1}\textbf{0.095}    \\
\bottomrule
\end{tabular}
\vspace{-3pt}
\caption{\textbf{Pose estimation results on the RE10K \cite{zhou2018stereo} and ACID \cite{infinite_nature_2020}.} We evaluate the pose estimation accuracy of our method with different numbers of input views. Our method outperforms other baseline methods in terms of pose accuracy.}
\label{tab:ate_comparison}
\vspace{-3pt}
\end{table}

\begin{figure}
  \centering
  \includegraphics[width=\columnwidth]{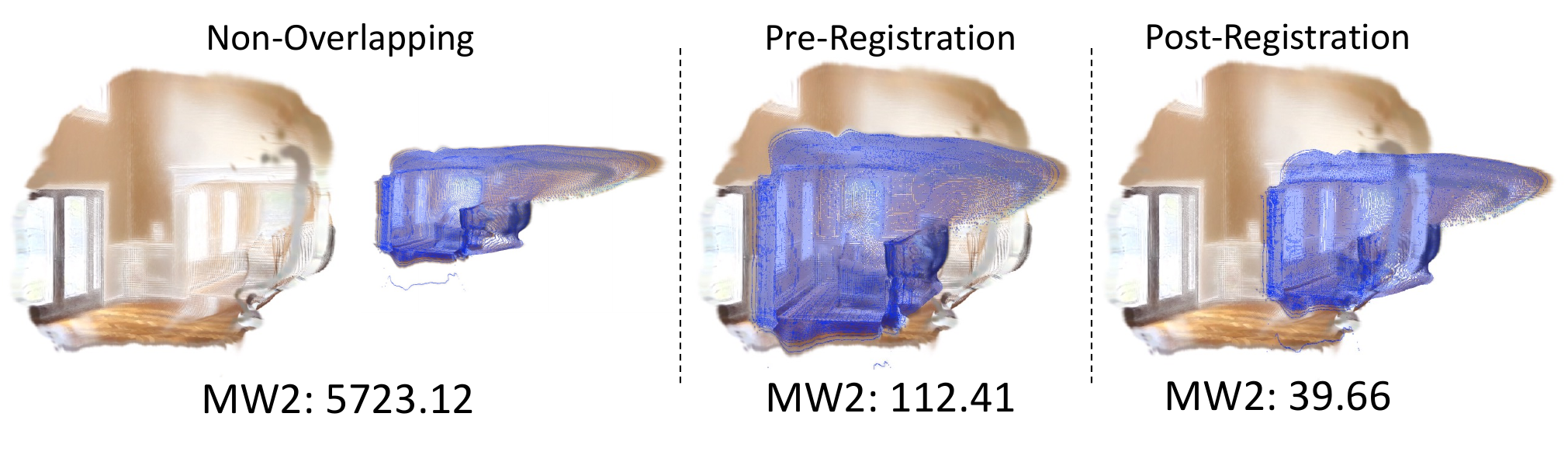}
  \vspace{-5pt}
  \caption{$\text{MW}_2$ distances effectively quantify alignment levels between sets of 3D Gaussians under various conditions. Notably, the rightmost case aligns with the correct position.}
  \label{fig:re10_mw2}
    \vspace{-15pt}
\end{figure}

\begin{figure}
  \centering
  \includegraphics[width=0.46\textwidth]{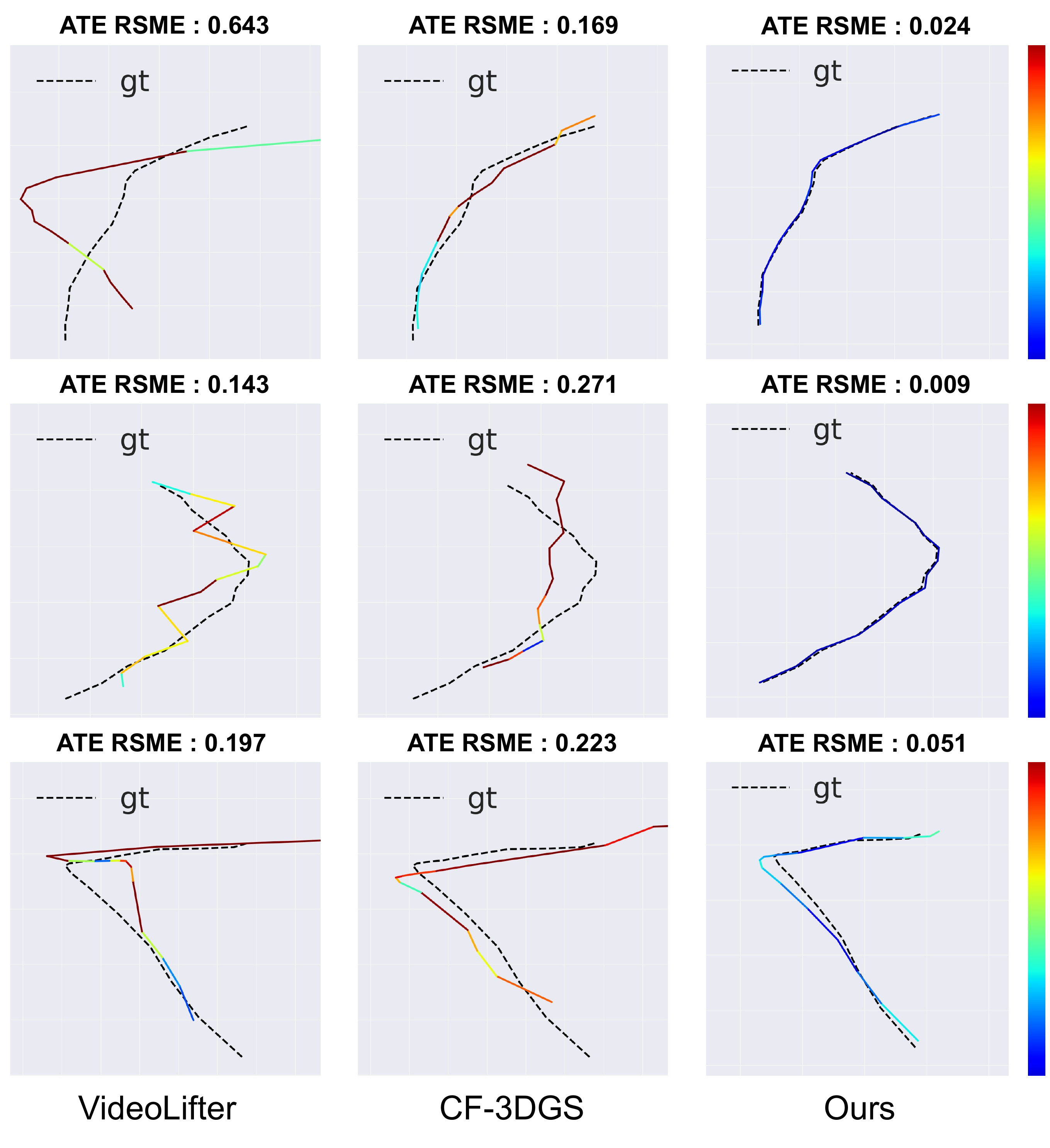}
  \vspace{-10pt}
  \caption{\textbf{Trajectory Comparison on the RE10K \cite{zhou2018stereo}.} Our method and the baseline are under 16-view input. Our method achieves higher pose estimation accuracy than other unposed methods and is applicable to various scenes and camera motions.}
  \label{fig:re10_traj}
    \vspace{-10pt}
\end{figure}

\section{Experiment}
\subsection{Experiment Setup}
\noindent\textbf{Datasets.} To evaluate the effectiveness of our method, we conducted experiments on the RE10K \cite{zhou2018stereo} and ACID \cite{infinite_nature_2020} datasets. The RE10K dataset includes indoor and outdoor scene videos, while ACID consists mainly of aerial shots of natural landscapes captured by drones. Both provide camera poses and intrinsic parameters. Following the setup in \cite{ye2024no}, we use the test sets of each dataset for evaluation.

For the unposed sparse views reconstruction task, the number of views we reconstructed are 2, 8, 16, and 32. To simulate sparse input, both training and testing views are equidistantly sampled from the videos. For 2-view scenarios, we sample every 40 frames for videos with significant motion and every 60 frames for scenes with less motion. For scenarios with 8, 16, and 32 views, training views are equidistantly sampled throughout the entire video. The test set includes all frames not used for training.

\noindent\textbf{Evaluation Metrics.} To evaluate novel view synthesis (NVS), we use PSNR, SSIM, and LPIPS as metrics. For pose estimation evaluation, we use ATE RMSE as a metric. For 3DGS registration evaluation, we use the $\text{MW}_2$ distance. As illustrated in \cref{fig:re10_mw2}, the proposed $\text{MW}_2$ distance precisely quantifies the proximity between two GMMs.

\noindent\textbf{Baselines.} We compare our method with methods for unposed reconstruction in the NVS task, including: Colmap \cite{schoenberger2016mvs, schoenberger2016sfm}, NoPoSplat \cite{ye2024no}, NoPe-NeRF \cite{bian2023nope}, VideoLifter \cite{cong2025videolifter}, CF-3DGS \cite{fu2024colmap}, MARSt3R \cite{leroy2024groundingimagematching3d}, and Splatt3R \cite{smart2024splatt3rzeroshotgaussiansplatting}.

\noindent\textbf{Implementation Details.} The hardware used in our experiments is the NVIDIA A6000. Our method is implemented using PyTorch, with NoPoSplat \cite{ye2024no} as the backbone. In the pose estimation of training frames and the scale estimation of sub Gaussians, we perform joint optimization. After completing registration and optimization for all frames, we perform global refinement to further refine the scene.

\subsection{Experimental Results and Analysis}

\noindent\textbf{Novel View Synthesis:} As shown in \cref{tab:re10k_comparison}, \cref{tab:acid_comparison}, \cref{fig:re10k_visual_diff} and \cref{fig:acid_visual_diff}, our method significantly outperforms other unposed reconstruction methods in terms of PSNR and SSIM. NoPe-NeRF \cite{bian2023nope} fails to converge; VideoLifter \cite{cong2025videolifter} produces distorted renderings under sparse views; and CF-3DGS \cite{fu2024colmap} suffers from artifacts due to inadequate detail capture. For LPIPS, we generally lead, though we occasionally fall short in some cases, due to noise introduced by global refinement when improving PSNR.

\noindent\textbf{Pose Estimation}: Our method can also be applied to pose estimation. As shown in \cref{tab:ate_comparison} and \cref{fig:re10_traj}. We conduct experiments on the RE10K \cite{zhou2018stereo} and ACID \cite{infinite_nature_2020} datasets under 8, 16, and 32-view input conditions. The poses estimated by the baseline methods were aligned with the ground truth (GT) poses for comparison. \Cref{tab:ate_comparison} presents the performance of our method. Compared to other unposed methods, our method demonstrates a more pronounced performance gap, especially in sparse view conditions.

\subsection{Ablation Studies}
\noindent\textbf{Ablation Study on Loss Function:} In this section, we investigate the 3DGS joint optimization loss function described in \cref{sec:joint_registration}. To validate the performance of our designed loss function, we conduct experiments on the RE10K \cite{zhou2018stereo} dataset by testing the results when each individual loss term is omitted. The input is set to 16 views, and the evaluation metrics used for comparison are ATE, PSNR, SSIM, LPIPS, and $\text{MW}_2$. To facilitate the comparison of the $\text{MW}_2$ loss, we normalize its values to a range of 0 to 100, representing the baseline for convergence.

As shown in \cref{fig:re10_mw2}, the $\text{MW}_2$ distance measures the closeness of the Gaussian scene structure distribution. Experiments in \cref{tab:ablations_loss} demonstrate that the $\text{MW}_2$ loss supports coarse alignment and pose estimation but may lead to local minima and misalignment when used alone. Photometric loss is essential for refining registration and improving NVS, yet it may cause submaps to converge to separate spatial regions. Depth-consistency loss stabilizes pose and geometry but fails to converge in isolation. These results underscore the necessity of jointly optimizing all loss terms for accurate registration.


\noindent\textbf{Ablation Study on Key Module:} The key module in our approach is the joint 3DGS registration. We perform experiments following the same setup as in the previous experiments. As shown in \cref{tab:ablations_module}, when the 3DGS joint registration module is removed, there is a significant decline in scene reconstruction and pose estimation accuracy, indicating the critical role of this module in accurate pose estimation and 3DGS registration.

\begin{table}[]
\small
\centering
\setlength{\tabcolsep}{3pt} 
\begin{tabular}{l|ccccc}
\toprule
               & ATE↓           & PSNR↑          & SSIM↑         & LPIPS↓        & $\text{MW}_2$↓                \\ \midrule
w/o Photo & 1.184          & 16.06          & 0.52          & 0.44          & 58.8          \\
w/o Depth & 0.160          & 20.97          & 0.72          & 0.29          & 57.8          \\
w/o MW2        & 1.151          & 19.41          & 0.67          & 0.31          & 67.7          \\ \midrule
\textbf{RegGS (Ours)}           & \textbf{0.098} & \textbf{23.09} & \textbf{0.79} & \textbf{0.23} & \textbf{56.5} \\
\bottomrule
\end{tabular}
\vspace{-7pt}
\caption{\textbf{Ablations on Loss Functions.} The performance of our method degrades when any loss term is removed, demonstrating the effectiveness of the loss functions we employ.}
\label{tab:ablations_loss}
\vspace{-8pt}
\end{table}

\begin{table}[]
\small
\centering
\setlength{\tabcolsep}{3pt}
\begin{tabular}{l|ccccc}
\toprule
                            & ATE↓           & PSNR↑          & SSIM↑         & LPIPS↓        & $\text{MW}_2$↓                \\ \midrule
w/o JR & 1.164          & 11.41          & 0.34          & 0.60          & 100.0          \\ \midrule
\textbf{RegGS (Ours)}                        & \textbf{0.098} & \textbf{23.09} & \textbf{0.79} & \textbf{0.23} & \textbf{56.5} \\
\bottomrule
\end{tabular}
\vspace{-7pt}
\caption{\textbf{Ablations on key Modules.} The results show that precise pose estimation and 3DGS registration depend on the 3DGS joint registration (JR) module.}
\label{tab:ablations_module}
\vspace{-8pt}
\end{table}

\subsection{Limitations}
Our method is influenced by the performance of feed-forward Gaussians; poor quality generation by these models can lead to registration and fusion failures. Additionally, the training time increases significantly with more input views due to the $\text{MW}_2$ distance, indicating the need for further optimization. In cases of large inter-frame motion, the registration process may also fail to converge.

\section{Conclusion}

This paper presents RegGS, an incremental 3D Gaussian reconstruction framework for unposed sparse view settings. We constructed a GMM alignment metric in $\mathrm{Sim}(3)$ space based on the optimal transport $\text{MW}_2$ distance, and efficiently computed the $\text{MW}_2$ distance using the entropy-regularized Sinkhorn algorithm, thereby circumventing the infinite-dimensional optimization problem. By jointly optimizing $\text{MW}_2$, photometric, and depth-consistency losses, RegGS achieves progressive coarse-to-fine registration of both camera poses and scene structure. Experiments on RE10K and ACID demonstrate superior pose estimation and novel view synthesis compared to prior methods, highlighting RegGS’s potential for real-world applications.

\clearpage
\section*{Acknowledgment}
This research is supported by the National Natural Science Foundation of China (No. 62406267), Guangzhou-HKUST(GZ) Joint Funding Program (Grant No.2025A03J3956 \& Grant No.2023A03J0008), the Guangzhou Municipal Science and Technology Project (No. 2025A04J4070), and the Guangzhou Municipal Education Project (No. 2024312122).
{
    \small
    \bibliographystyle{ieeenat_fullname}
    \bibliography{main}
}
\clearpage
\setcounter{page}{1}
\maketitlesupplementary

\section{Entropy-Regularized Sinkhorn $W2$ Distance Gradient Consistency Proof}

The entropy-regularized Wasserstein distance $W_{2,\epsilon}^2(G_A, T(G_B))$, where $\epsilon > 0$ is a regularization parameter, provides a computationally feasible approach to the infinite-dimensional optimization problem inherent in calculating the exact Wasserstein distance $W_2^2$. As $\epsilon \rightarrow 0$, the gradient $\nabla_\xi W_{2,\epsilon}^2$ converges to the subgradient set of the exact Wasserstein distance, denoted as $\partial W_2^2(\xi)$:
\begin{equation}
\lim_{\epsilon \rightarrow 0} \nabla_\xi W_{2,\epsilon}^2 \in \partial W_2^2(\xi).
\end{equation}

\textbf{Foundation of $\Gamma$-Convergence.} According to optimal transport theory \cite{Sinkhorn}, the entropy-regularized Wasserstein distance satisfies $\Gamma$-convergence: for any probability distributions $G_A$ and $G_B$, as the regularization parameter $\epsilon$ approaches zero,
\begin{equation}
W_{2,\epsilon}^2(G_A, G_B) \xrightarrow{\Gamma} W_2^2(G_A, G_B),
\end{equation}
where $\Gamma$-convergence ensures that the sequence of minima of the regularized problem converges to the optimum of the original problem. Specifically, for a parametrized transformation $T(\xi)$, the minimization of the entropy-regularized objective function $W_{2,\epsilon}^2$ approximates the non-regularized objective $W_2^2$ in the limit.

\subsection{Gradient Expression Derivation.} 
The entropy-regularized Wasserstein distance is defined as:
\begin{equation}
W_{2,\epsilon}^2 = \min_{\pi \in \Pi(w_A, w_B)} \sum_{i,k} \pi_{ik} C_{ik}(\xi) + \epsilon \sum_{i,k} \pi_{ik} \log \pi_{ik},
\end{equation}
where $C_{ik}(\xi) = \| \mu_i^A - \mu_k^{B\prime} \|^2 + \operatorname{Tr}(\Sigma_i^A + \Sigma_k^{B\prime} - 2 (\Sigma_i^A \Sigma_k^{B\prime})^{1/2})$ depends on the transformation parameters $\xi$. Using the implicit function theorem \cite{doi:10.1137/1.9781611971309}, the gradient of the regularized problem can be expressed as:
\begin{equation}
\nabla_\xi W_{2,\epsilon}^2 = \sum_{i,k} \pi_{ik}^*(\epsilon) \nabla_\xi C_{ik}(\xi),
\end{equation}
where $\pi_{ik}^*(\epsilon)$ is the optimal transport plan under entropy regularization. This expression indicates that the gradient is constituted by a weighted average of the cost function gradients under the transport plan.

\subsection{Convergence of the Transport Plan.} 
As $\epsilon \rightarrow 0$, the influence of the entropy regularization term $\epsilon \sum \pi_{ik} \log \pi_{ik}$ diminishes. According to $\Gamma$-convergence, any limit point of the regularized transport plan $\pi_{ik}^*(\epsilon)$ is an optimal solution of the original Wasserstein problem, i.e.,
\begin{equation}
\lim_{\epsilon \rightarrow 0} \pi_{ik}^*(\epsilon) = \pi_{ik}^* \in \arg \min_\pi \sum_{i,k} \pi_{ik} C_{ik}(\xi).
\end{equation}
Since multiple optimal transport plans may exist (e.g., multiple paths with the same minimum cost), $\pi^*$ belongs to a set of optimal solutions $\Pi^*$.

\subsection{Construction of the Subgradient Set}

For a nonsmooth convex function $W_2^2$, its Clarke subgradient is defined as:
\begin{equation}
\partial W_2^2(\xi) = \left\{ \sum_{i,k} \pi_{ik}^* \nabla_\xi C_{ik}(\xi) \mid \pi^* \in \Pi^* \right\}.
\end{equation}
As $\epsilon \rightarrow 0$, the limit points of the regularized gradient $\nabla_\xi W_{2,\epsilon}^2 = \sum_{i,k} \pi_{ik}^* (\epsilon) \nabla_\xi C_{ik}(\xi)$ are determined by the convergence of $\pi_{ik}^*(\epsilon)$. Thus,
\begin{equation}
\lim_{\epsilon \rightarrow 0} \nabla_\xi W_{2,\epsilon}^2 = \sum_{i,k} \pi_{ik}^* \nabla_\xi C_{ik}(\xi) \in \partial W_2^2(\xi),
\end{equation}
indicating that the regularized gradient converges to an element of the subgradient set.

Although \(W_2^2\) may be nonconvex with respect to \(\xi\), it satisfies local Lipschitz continuity on any compact set \cite{santambrogio2015optimal}, ensuring the existence of subgradients.

If the original problem has a unique optimal transport plan \(\pi^*\), then the subgradient reduces to a singleton and the gradient convergence path is unique; otherwise, convergence is toward a specific direction within the subgradient set.

The entropy-regularized gradient $\nabla_\xi W_{2,\epsilon}^2$ asymptotically approaches the exact Wasserstein subgradient direction as $\epsilon$ approaches zero. This property theoretically supports the hierarchical optimization strategy of gradually reducing $\epsilon$ in our methodology: initially leveraging the smoothness of the regularization term to avoid local minima and eventually converging towards the direction of the exact Wasserstein distance, thus achieving robust global distribution alignment.

\section{Sinkhorn Algorithm Complexity}
The main map and the submap contain $M$ and $N$ Gaussian gradients, respectively. Initially, the first step of the Sinkhorn algorithm involves constructing a kernel matrix $K \in \mathbb{R}^{M \times N}$, whose elements are given by
\begin{equation}
K_{ik} = \exp\left(-\frac{C_{ik}}{\epsilon}\right),
\end{equation}
where $C_{ik}$ represents the 2-Wasserstein cost for the Gaussian pair $(N_i^A,\,N_k^{B'})$. Calculating each $C_{ik}$ includes two parts: one is the distance between means $\|\mu_i^A - \mu_k^{B'}\|^2$, which has a complexity of $O(1)$; the second is the covariance term
\begin{equation}
\operatorname{Tr}\left(\Sigma_i^A + \Sigma_k^{B'} - 2\left(\Sigma_i^A\,\Sigma_k^{B'}\right)^{1/2}\right),
\end{equation}
where the square root of the covariance matrix is usually implemented via Cholesky decomposition, with a single conjunction complexity of $O(d^3)$ (for a three-dimensional space $d=3$), but since all Gaussian covariances can be precomputed, this can be considered $O(1)$ in the context of $C_{ik}$ computation. Thus, the overall complexity of constructing the kernel matrix $K$ is $O(MN)$.

Next, during the iteration phase, the Sinkhorn algorithm manages $u \in \mathbb{R}^{M}$ and $v \in \mathbb{R}^{N}$ through alternating updates to satisfy the marginal constraints, with the update formulas
\begin{equation}
u^{(t)} = \frac{w_A}{K\,v^{(t-1)}}, \qquad v^{(t)} = \frac{w_B}{K^\top\,u^{(t)}}.
\end{equation}
Here, the complexity of multiplying the matrix with the support (i.e., computing $K\,v^{(t-1)}$ and $K^\top\,u^{(t)}$) incurs $O(MN)$, while the element-wise division to update $u^{(t)}$ and $v^{(t)}$ has complexities of $O(M)$ and $O(N)$, respectively, which are negligible compared to the previous step. Therefore, each iteration's computational complexity is $O(MN)$.

After $T$ iterations, the total time complexity is the kernel matrix initialization $O(MN)$ plus $T \cdot O(MN)$, which is
\begin{equation}
O(MN) + T\cdot O(MN) = O(TMN).
\end{equation}
In practice, due to the introduction of the entropy regularization term, which significantly speeds up convergence, according to \cite{Sinkhorn}, the Sinkhorn algorithm typically converges within $T \leq 50$ iterations to a relative parameter $\delta < 10^{-3}$, a characteristic that has been verified in multiple optimal transport libraries such as POT \cite{flamary2021pot}. The detailed procedure is described in the following Algorithm 1.

\begin{algorithm}
\caption{Entropy-Regularized Optimal Transport \(MW_2\) Distance}
\begin{algorithmic}[1]
\Require 
  \begin{itemize}
    \item Gaussian components: \(\{\mathcal{N}(\mu_i^A, \Sigma_i^A)\}_{i=1}^{M}\) and \(\{\mathcal{N}(\mu_k^{B'}, \Sigma_k^{B'})\}_{k=1}^{N}\).
    \item Marginal weights: \(w^A \in \mathbb{R}^{M}\) and \(w^B \in \mathbb{R}^{N}\).
    \item Regularization parameter: \(\epsilon > 0\).
    \item Maximum iterations: \(T\).
  \end{itemize}
\Ensure 
  \begin{itemize}
    \item Transport plan \(\pi^* \in \mathbb{R}^{M \times N}\).
    \item Entropy-regularized transport cost \(W_{2,\epsilon}^2\).
  \end{itemize}

\State \textbf{Step 1: Compute Cost Matrix \(C\)}
\For{\(i = 1\) to \(M\)}
    \For{\(k = 1\) to \(N\)}
        \State \(C_{ik} \gets \|\mu_i^A - \mu_k^{B'}\|^2 + \operatorname{Tr}\Bigl(\Sigma_i^A + \Sigma_k^{B'} - 2\bigl(\Sigma_i^A \Sigma_k^{B'}\bigr)^{1/2}\Bigr)\)
    \EndFor
\EndFor

\State \textbf{Step 2: Compute Kernel Matrix \(K\)}
\For{\(i = 1\) to \(M\)}
    \For{\(k = 1\) to \(N\)}
        \State \(K_{ik} \gets \exp\Bigl(-\frac{C_{ik}}{\epsilon}\Bigr)\)
    \EndFor
\EndFor

\State \textbf{Step 3: Initialize scaling vectors}
\State \(u \gets \mathbf{1}_M\) \Comment{\(\mathbf{1}_M\) denotes a vector of ones with length \(M\)}
\State \(v \gets \mathbf{1}_N\) \Comment{\(\mathbf{1}_N\) denotes a vector of ones with length \(N\)}

\State \textbf{Step 4: Perform Sinkhorn Iterations}
\For{\(t = 1\) to \(T\)}
    \State \(u \gets w^A \oslash (K\,v)\) \Comment{\(\oslash\) denotes element-wise division}
    \State \(v \gets w^B \oslash (K^\top u)\)
\EndFor

\State \textbf{Step 5: Compute the Transport Plan}
\State \(\pi^* \gets \operatorname{diag}(u) \, K \, \operatorname{diag}(v)\)

\State \textbf{Step 6: Compute the Entropy-Regularized Transport Cost}
\State \(W_{2,\epsilon}^2 \gets \sum_{i=1}^{M} \sum_{k=1}^{N} \pi_{ik}^*\,C_{ik}\)

\State \Return \(\pi^*,\; W_{2,\epsilon}^2\)
\end{algorithmic}
\end{algorithm}

\begin{figure*}
\centering
\begin{tabular}{l|ccc|ccc|ccc}
\toprule
\multirow{2}{*}{\textbf{Method}} & \multicolumn{3}{c}{2×} & \multicolumn{3}{c}{16×} & \multicolumn{3}{c}{64×}            \\ \cmidrule(lr){2-4} \cmidrule(lr){5-7} \cmidrule(lr){8-10} 
            & PSNR↑  & Time↓ & GPU(GB)↓ & PSNR↑  & Time↓ & GPU(GB)↓ & PSNR↑  & Time↓ & GPU(GB)↓ \\ \midrule
Splatt3R    & 13.951 & \textbf{20s} & 7.3 & -      & -     & -      & -      & -     & -      \\
NoPoSplat   & 23.247 & 22s & 3.9 & -      & -     & -      & -      & -     & -      \\
DUSt3R*     & 18.484 & 259s & \textbf{3.5} & 24.714 & \textbf{22min} & 10.5 & OOM    & OOM   & OOM    \\
MASt3R*     & 16.036 & 283s & 3.7 & 24.249 & 23min & \textbf{4.5}  & \textbf{28.826} & \textbf{54min} & 41.3 \\
\midrule
\textbf{Ours} & \textbf{24.272} & 259s & 3.9 & \textbf{28.663} & 57min & 12.1 & 28.703 & 165min & \textbf{12.1} \\
\bottomrule
\end{tabular}
\vspace{-6pt}
\caption{\textbf{Additional quantitative comparison on RE10K showing runtime and memory usage across different input view counts.}}
\label{tab:sup_re10k_exp1}
\end{figure*}

Moreover, although the theoretical time complexity is $O(TMN)$, in engineering implementations, various strategies can reduce the sparsity factor, utilizing GPU sparsity for computing the kernel matrix $K$ and matrix multiplications; employing safe logarithmic domain computations (i.e., computing $\log K_{ik} = -C_{ik}/\epsilon$) to sparsify and reduce multiplication/division operations; and leveraging sparsity methods to speed up by building a sparse kernel matrix reducing the computational complexity to $O(S)$ (where $S \ll MN$ is the number of non-zero elements).

In summary, under entropy regularization, the Sinkhorn algorithm's time complexity is $O(TMN)$, where $T$ is the number of iterations (usually $T \leq 50$). In practical applications of 3D Gaussian map registration (e.g., $M, N \leq 10^5$), a single iteration takes about 10ms, and with GPU sparsification and sparsity priors, this method can achieve fast processing of large-scale 3D Gaussian map registration problems.

\section{Additional Experimental Results}
\begin{figure}
  \centering
  \includegraphics[width=0.45\textwidth]{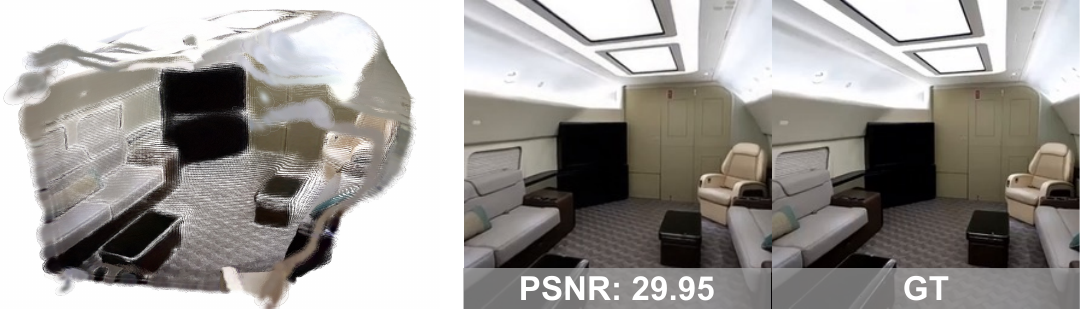}
  \vspace{-8pt}
  \caption{Generalization results using a real video. Four frames were uniformly sampled and used for sparse reconstruction to demonstrate the method's applicability to real-world scenarios.}
  \label{fig:real}
    \vspace{-8pt}
\end{figure}

\Cref{fig:real} illustrates the generalization of our method on real video data, where we uniformly sampled four frames from a video and used a pretrained feed-forward Gaussian model to extract local 3D Gaussian representations. These representations were registered and fused by the RegGS method into a consistent 3D Gaussian scene. Results confirm the method's effectiveness in achieving precise camera localization and scene alignment even with sparse viewpoints, thus generating high-quality novel views suitable for real-world applications.

\begin{figure}
  \centering
  \includegraphics[width=0.45\textwidth]{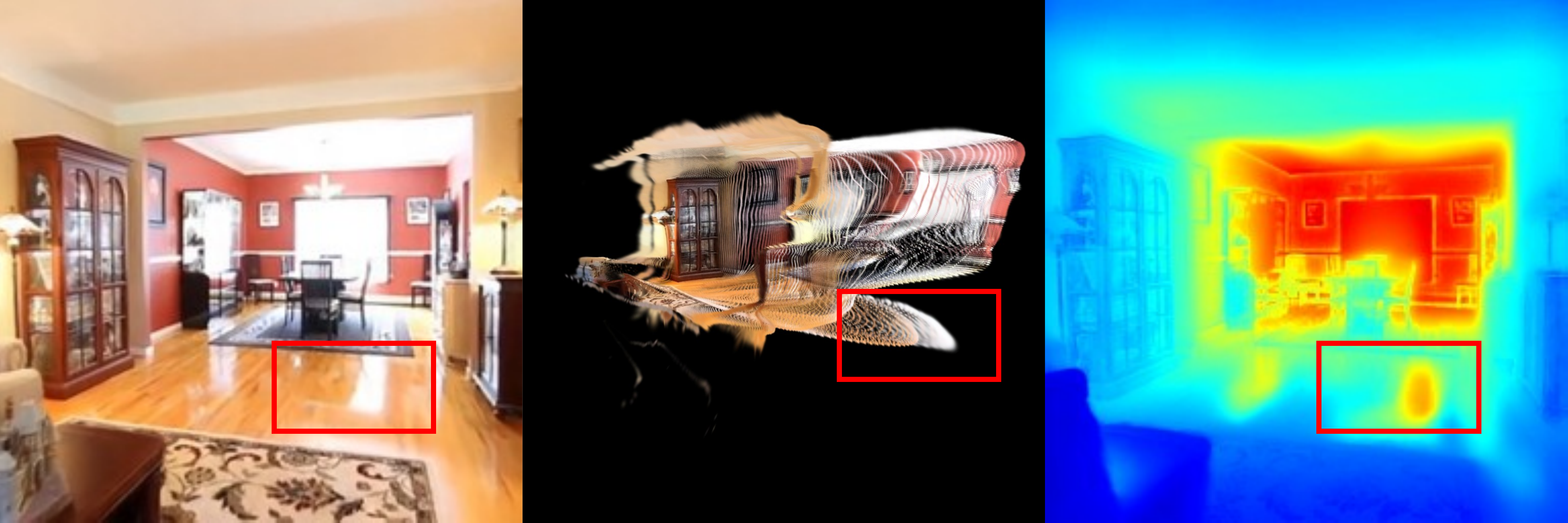}
    \vspace{-8pt}
  \caption{Sub gaussians from NoPoSplat. These sub Gaussians generated by NoPoSplat indicate that, in certain scenes, the Gaussians produced by NoPoSplat exhibit abnormalities in their spatial structure.}
  \label{fig:limitations1}
    \vspace{-8pt}
\end{figure}

\begin{figure}
  \centering
  \includegraphics[width=0.45\textwidth]{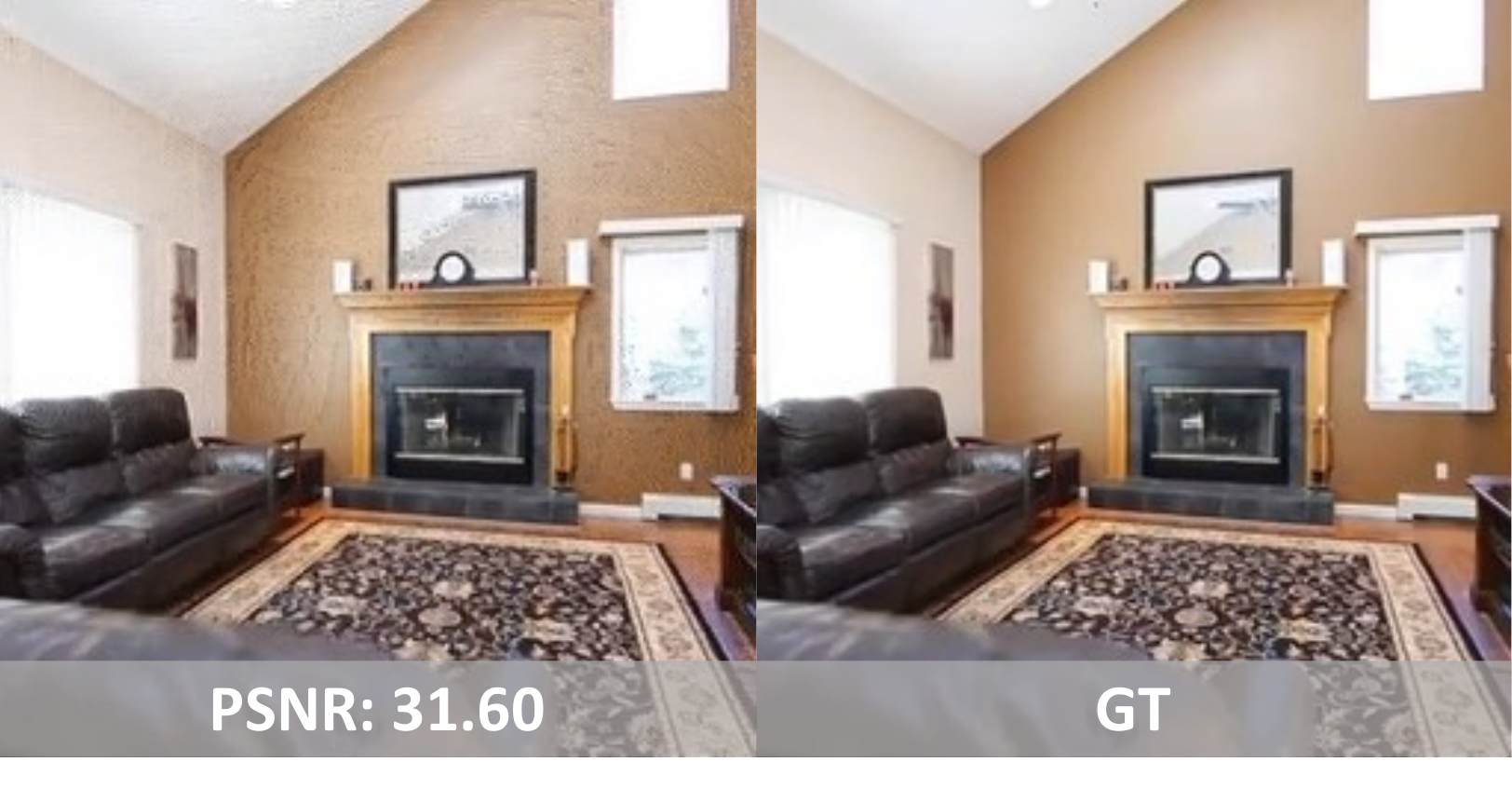}
    \vspace{-8pt}
  \caption{Visual Comparison. It is noticeable that there are many prominent noise points on the wall. Our method, in certain scenes, may produce high PSNR values, but visually, there are clearly visible noise artifacts.}
  \label{fig:limitations2}
    \vspace{-8pt}
\end{figure}

We evaluate runtime and memory across sampling ratios from 2x to 64x using 200-frame sequences. As shown in Tab. \ref{tab:sup_re10k_exp1}, RegGS maintains controlled memory usage across all input settings. In contrast, NoPoSplat and Splatt3R are limited to two-view inputs, while DUSt3R and MASt3R exhibit exponential growth in Gaussian count, frequently resulting in out-of-memory failures. This demonstrates the scalability of RegGS under sparse view conditions. At 64x, as view coverage becomes denser, the reconstruction bottleneck shifts from view sparsity to the capacity of the 3DGS representation. RegGS achieves comparable reconstruction quality to MASt3R with significantly lower memory consumption. Further optimization of $\text{MW}_2$ computation remains a direction for future work.

\section{Additional Limitations}

\Cref{fig:limitations1} demonstrates that NoPoSplat generates suboptimal Gaussians in certain scenes. In the depicted scenario, NoPoSplat struggles to accurately estimate the depth information of the reflective surface, causing the gaussians to fail at capturing the spatial geometry effectively. RegGS relies on the quality of the Gaussian model generated by the upstream model, and abnormal gaussians introduced during scene fusion can lead to errors.

\Cref{fig:limitations2} shows that, in certain scenes, while our method achieves high PSNR values, there are noticeable noise artifacts. These noise points are likely introduced during the refinement stage or could be a result of the low image resolution used in our quantitative evaluation. In future work, we will attempt to address this issue.

\end{document}